\documentclass[accepted,specialissue]{melba}

\usepackage{amsmath,amsfonts}\usepackage{graphicx}
\usepackage{caption}
\usepackage{subcaption}
\usepackage{lscape}
\usepackage{mathrsfs}
\usepackage{wrapfig} 
\usepackage{tabularx}

\usepackage{fancyhdr}
\usepackage{afterpage}
\usepackage{bbm}
\usepackage[table]{xcolor}
\usepackage[export]{adjustbox}
\newif\iftopic
\topicfalse 

\melbaid{2025:027}  
\doi{10.59275/j.melba.2025-ae66}
\melbaauthors{Dominique et al.}  
\email{dominique.b@northeastern.edu}
\volume{3}
\firstpageno{618}  
\melbayear{2025}  
\datesubmitted{04/2025}  
\datepublished{10/2025}  

\melbaspecialissue{Special issue on Fairness of AI in Medical Imaging (FAIMI)}
\melbaspecialissueeditors{Veronika Cheplygina, Aasa Feragen, Andrew King, Ben Glocker, Enzo Ferrante, Eike Petersen, Esther Puyol-Antón, Melanie Ganz-Benjaminsen}

\ShortHeadings{MELBA Journal Sample Article}{Dominique et al.}

\title{On the Role of Calibration in Algorithmic Fairness Benchmarking for Skin Cancer Detection}
\author{
	\firstname Brandon \surname Dominique\aff{1}\orcid{0009-0002-5163-6538},
	\firstname Prudence \surname Lam\aff{1}\orcid{0009-0000-9753-8234},
        \firstname Nicholas \surname Kurtansky\aff{2}\orcid{0000-0002-6745-0386},
	\firstname Jochen \surname Weber\aff{2}\orcid{000-0002-2397-4954},
        \firstname Kivanc  \surname Kose\aff{2}\orcid{0000-0003-3185-2639},
	\firstname Veronica \surname Rotemberg\aff{2}\orcid{0000-0003-0639-2677},
        \firstname Jennifer \surname Dy\aff{1}\orcid{0000-1111-2222-3333}
}
\affiliations{
	\num 1 \addr Northeastern University, Boston, MA, USA 02115 \\
	\num 2 \addr Memorial Sloan Kettering Cancer Center, New York, NY, USA 10065 \\
}

\abstract{
Artificial Intelligence (AI) models have demonstrated expert-level performance in melanoma detection, yet their clinical adoption is hindered by performance disparities across demographic subgroups such as gender, race, and age. Previous efforts to benchmark the performance of AI models have primarily focused on assessing model performance using group fairness metrics that rely on the Area Under the Receiver Operating Characteristic curve (AUROC), which does not provide insights into a model's ability to provide accurate estimates. In line with clinical assessments, this paper addresses this gap by incorporating calibration as a complementary benchmarking metric to AUROC-based fairness metrics. Calibration evaluates the alignment between predicted probabilities and observed event rates, offering deeper insights into subgroup biases. \textcolor{black}{We assess the performance of the leading skin cancer detection algorithm of the ISIC 2020 Challenge on the ISIC 2020 Challenge dataset and the PROVE-AI dataset, and compare it with the second- and third-place models, focusing on subgroups defined by sex, race (Fitzpatrick Skin Tone), and age.} Our findings reveal that while existing models enhance discriminative accuracy, they often over-diagnose risk and exhibit calibration issues when applied to new datasets. This study underscores the necessity for comprehensive model auditing strategies and extensive metadata collection to achieve equitable AI-driven healthcare solutions. All code is publicly available at \url{https://github.com/bdominique/testing_strong_calibration}.}


\usepackage{float}
\begin{document}

\twocolumn[\maketitle]

\section{Introduction}\label{sec:Intro}



Artificial Intelligence (AI) models have achieved expert-level performance in melanoma detection \citep{isic2020}. However, their clinical adoption remains limited due to performance disparities across demographic subgroups, including gender, race, and age ~\citep{Feng2023IsTM}. These disparities are further compounded by differences in melanoma incidence and tumor presentation across the various subgroups~\citep{Shao2022, Marchetti2021}. Addressing these challenges requires comprehensive model auditing strategies that can identify and evaluate subgroup biases before deployment. 




One emerging field of AI auditing has been Fairness Benchmarking, characterized by the development of rigorous benchmarking protocols for model assessment under different fairness settings~\citep{FB1, FB2, FB3}. 
Recent fairness benchmarking methods on clinical data have primarily emphasized \emph{Group Fairness}, which ensures that AI models perform equitably across different demographics. These works utilize a suite of metrics related to the area under the receiver operating characteristic curve (AUROC) to check for balanced predictive ability across subgroups~\citep{zong2023medfairbenchmarkingfairnessmedical, jin2024fairmedfmfairnessbenchmarkingmedical}.
However, while AUROC captures a model's discriminative power, it does not provide insight into whether a model systematically \emph{over}- or \emph{under}- estimates subgroup risk ~\citep{Huang2020ATO}. 
Consider, for instance, a model that predicts the success of vitro fertilization (IVF) -- a procedure with variable efficacy designed to assist with conception. Even if such a model accurately classifies successful versus unsuccessful treatments, its clinical utility diminishes if it consistently mispredicts the likelihood of a live birth. Underestimation can increase psychological distress for patients, which in turn impacts specific stages of the IVF process~\citep{Zanettoullis2024}.
Conversely, overestimation can foster false hope and result in the oversight of critical pre-procedure interventions ~\citep{inovifertilitySuccessRates}. To address this limitation, we include an additional metric focused on \emph{calibration}, which quantifies the degree to which a model's average predicted probabilities align with observed event rates ~\citep{Huang2020ATO}.

    
Selecting an appropriate calibration metric is critical, as different approaches exhibit different sensitivities to subgroup sizes ~\citep{Jankov2020, Zhang2021}. Moreover, the choice of relevant subgroups can impose additional biases onto the calibration framework. 
In order to address these shortcomings, we incorporate a calibration method based on an adaptive-score Cumulative Sum (CUSUM) test as one of our benchmarking tools ~\citep{Feng2023IsTM}. This method enables efficient detection of miscalibration across all subgroups present in the audit dataset, without being restricted to a predefined or limited set of candidate subgroups. 


     
This study evaluates the performance of a state of the art skin cancer detection classification algorithm, winner of the ISIC 2020 Challenge (hereafter ADAE)~\citep{ha2020identifyingmelanomaimagesusing}, in comparison to the second- and third-place models (hereafter 2nd Place and 3rd Place, respectively)~\citep{githubKagglemelanomadocumentationpdfMaster,kaggleSIIMISICMelanoma} when applied to two datasets of varying sizes with a specific focus on subgroups defined by \emph{sex}, \emph{race} (represented in this study by Fitzpatrick Skin Tone; \emph{FST} hereafter), and \emph{age} - specifically, one dataset contains each patient's sex and age, while the other contains all 3. Moreover, we account for intersectionality (i.e., patients spanning multiple subgroups) when evaluating for both AUROC and calibration. \textcolor{black}{The intersectional AUROC analysis is performed by comparing the discrimination of the model between combinations of demographic subgroups, using the DeLong test to assess statistically significant differences in AUROC between and within these combinations. Intersectional calibration analysis is done by applying a score-based CUSUM test and Variable Importance plots (hereafter \emph{VI plots}) to detect and explain miscalibration across combinations of demographic subgroups without requiring predefined subgroup lists.}
Our experimental results demonstrate that existing models perform comparably to the baseline when applied to datasets with previously unobserved risk factors, a finding that corroborates other benchmarking studies in this domain. Additionally, through our CUSUM calibration test we show that the inclusion or exclusion of patient risk factors results in variations in the model's calibration, such as certain risk factors being well calibrated when included and not calibrated when excluded. 

Section~\ref{sec: Related Work} reviews related work, Section~\ref{sec:Methods} describes our methodology, Section~\ref{sec: Results} presents our empirical study and its results, and Section~\ref{sec: DC} summarizes these results.

\section{Related Work} \label{sec: Related Work}

\subsection{Fairness in Medical Imaging} 

A substantial body of literature focuses on defining fairness for algorithmic decision-making systems in healthcare \citep{Chin2023, Grote2022, Bre2022}.  
In this context, fairness is often delineated along two principle dimensions: \emph{group fairness} and \emph{individual fairness}.

Individual fairness seeks to guarantee that similar patients -- such as ones based on relevant clinical features -- receive similar predictions or treatment recommendations~\citep{dwork2011fairnessawareness}. It aims to uphold the principle of personalized equity at the cost of a pre-defined similarity metric between individuals, which can be highly nuanced and challenging to obtain~\citep{Jui2024}. Given these considerations, we present two frameworks -— group fairness and calibration -— that we believe are most suitable for decision-making systems in healthcare due to their greater flexibility and relaxed assumptions regarding data structure.

 Group fairness aims to ensure equitable outcomes across socially salient groups, such as those defined by race, gender, or socioeconomic status, by enforcing independence between a model's predictions and the sensitive attribute~\citep{hutchinson, calders, feldman}. 
 Group fairness can be measured using many similarity metrics, such as the classification or misclassification rate of a predictive model amongst different subgroups~\citep{kusner2018counterfactualfairness,eo}; additionally, some definitions of group fairness do not focus on one metric but are designed to allow for a metric of the user's choosing~\citep{lahoti2020fairnessdemographicsadversariallyreweighted}. However, efforts to enforce group fairness can introduce trade-offs with model accuracy, particularly when there are underlying disparities in data representation or label quality. In clinical contexts, such trade-offs carry the risk of violating core bioethical principles such as non-maleficence and beneficence~\citep{beauchamp, zafar}, by exacerbating disparities in care or introducing harm to less-marginalized populations. 
 Additionally, group fairness metrics are often grounded in statistical measures, which not only impose inherent limitations but can also lead to mutual incompatibilities among fairness criteria~\citep{kearns, kleinberg}. These factors make the choice of group fairness metric(s) imperative, as it must be done in a way that does not exacerbate the problems that are trying to be mitigated. Motivated by standard model evaluation techniques from clinical contexts~\citep{Hong2023,Alba2017}, we expand our benchmark to  not only include group fairness metrics related to AUROC, but calibration as well (\textbf{Note:} in line with these techniques, we refer to our AUROC-based analysis as \emph{Model Discrimination} and our calibration-based analysis as \emph{Model Calibration} hereafter).

\subsection{Calibration}

\textbf{Calibration} is achieved in an AI model when the predicted probability of the model corresponds to the observed event rate. In other words, an AI model is calibrated when, given a population of people, the difference between its estimate and the true outcome is small. 
Different methods in the literature look to achieve this in different ways while also addressing issues such as variance in subgroup sizes and intersectional group memberships.
Specifically,~\cite{MultiCalib2018} created an influential method that asks for calibration not only on the entire population, but within specific subgroups as well. Other methods since then look for a similar type of strong guarantee~\citep{Multiaccuracy2019, luo2022localcalibrationmetricsrecalibration}. However, rather than checking for calibration amongst every subgroup, this method only checks over a predefined list of subgroups; additionally, these methods each require a high sample complexity in order to achieve any statistical guarantees over these subgroups. Feng et al. define their specific CUSUM test for Model Calibration in a way that allows an auditor to quickly check for miscalibration among all subgroups that appear in the audit dataset and is not limited to a small set of candidate subgroups. This test for strong calibration is also done in one pass of the data.
The VI plots that are included as an analysis tool in this method also help to identify when multiple subgroups could be the cause of miscalibration, addressing the issue of intersectional group memberships. These latter two reasons in particular are a significant advantage over methods that only allow for the test of a handful of subgroups at a time.
This method also does not require as many samples to guarantee strong calibration as its contemporaries, which typically ask for at least tens or even hundreds of thousands of samples.


\subsection{Fairness Benchmarking}
\textcolor{black}{There is growing interest in developing standardized benchmarks for assessing AI models in a trustworthy, reproducible, and cross-disciplinary way.} 
Current benchmarks emphasize the most widely-used fairness metrics and algorithms in literature ~\citep{FB1, FB2, FB3}.  In the medical domain specifically, recent benchmarking efforts have primarily used Model Discrimination to highlight disparities across models. For example, \cite{zong2023medfairbenchmarkingfairnessmedical} evaluated a diverse set of models across multiple data modalities (e.g., X-ray, dermatology images, MRI) and explored various model selection strategies using fairness metrics centered around AUROC. \cite{jin2024fairmedfmfairnessbenchmarkingmedical} expand upon their work to include the Dice Similarity Score, though its utility is limited to segmentation tasks in computer vision. \textcolor{black}{Our work builds upon these foundations by introducing a complementary perspective: evaluating the role of calibration metrics in fairness assessment.} While AUROC-based analysis is a helpful and intuitive way of comparing and understanding model performance, this type of analysis reveals no information about the calibration of these models. Calibration enhances the utility of a model's predictions by ensuring that the predicted probabilities are meaningful and actionable. Without a method for measuring and comparing the calibration of these models, catastrophic decisions could be made in any process that relies on these predictive values.

\section{Methodology}\label{sec:Methods}



\subsection{Datasets} 

The first dataset used in this experiment is the ISIC 2020 Challenge dataset~\citep{isic2020, Kurtansky2024, Rotemberg2021}. The Society for Imaging Informatics in Medicine and the International Skin Imaging Collaboration (SIIM-ISIC)'s 2020 Melanoma Classification Challenge is hosted on Kaggle and uses a convenience test set of 10,982 public dermoscopy images from six dermatology centers (Barcelona, Spain; New York, United States; Vienna, Austria; Sydney, Australia; Brisbane, Australia; Athens, Greece). The AUC scores for the challenge's final leaderboard were computed over a private, held-out portion of the dataset, while the AUC scores in this study were computed over the whole dataset. The full dataset includes 10,982 participants, with 43\% identifying as women and 51.6\% under the age of 50.

The second dataset is the PROVE-AI dataset~\citep{Marchetti2023}, comprised of 603 images prospectively collected at the Memorial Sloan Kettering Cancer Center (MSKCC), 95 of which are of melanoma. The dataset included 603 Participants (53.7\% Women,  49.1\% Under the age of 65). Due to the lack of samples from FSTs 4-6, we focus our evaluation on FSTs 1 and 2. However, our analysis is extendable to datasets with sufficient samples for these groups.

\subsection{Selected Methods}
In total, 3308 teams participated in the SIIM-ISIC 2020 challenge~\citep{Kurtansky2024}. To better understand the performance of the highest ranking approach on the experiments we designed, we also included the next 2 highest ranking approaches in our analysis. Detailed descriptions of the 3 highest ranking approaches are given below. To ensure that each method runs correctly, they were tested on the SIIM-ISIC 2020 data according to their respective publicly available scripts on GitHub\footnote{ADAE: \hyperlink{https://github.com/ISIC-Research/ADAE}{https://github.com/ISIC-Research/ADAE}} \footnote{2nd Place: \hyperlink{https://github.com/i-pan/kaggle-melanoma}{https://github.com/i-pan/kaggle-melanoma}} \footnote{3rd Place: \hyperlink{https://github.com/Masdevallia/3rd-place-kaggle-siim-isic-melanoma-classification}{https://github.com/Masdevallia/3rd-place-kaggle-siim-isic-melanoma-classification}}. \textcolor{black}{To test the discriminative ability and the calibration of each model when being used on a population different from the one it was trained on, we did not re-train the models on the PROVE-AI dataset and instead simply performed inference with the ISIC weights.}

\subsubsection{All Data are Ext (ADAE) (1st Place, ~\texorpdfstring{\cite{ha2020identifyingmelanomaimagesusing}}{ha2020identifyingmelanomaimagesusing})}
ADAE was the top-performing algorithm in the SIIM-ISIC 2020 challenge, achieving an overall AUROC of 0.9490. It employs an ensemble of eighteen Convolutional Neural Network (CNN) models—each trained across five validation folds, resulting in 90 sets of model weights. Sixteen of these models are based on the EfficientNet architecture, while two use ResNet. Additionally, four of the Sixteen EfficientNet models incorporate clinical metadata, including age, sex, anatomic site, and image size. Training sets from previous years of the challenge were included in the training process to mitigate class imbalances within the ISIC 2020 training set. Multiple image augmentations were also applied to prevent overfitting to the training data. The training involved a 5-fold cross-validation process, with the final model being an ensemble of these cross-validated models.

\subsubsection{2nd Place,~\texorpdfstring{\cite{githubKagglemelanomadocumentationpdfMaster}}{[2]}}
This method utilized an ensemble of fifteen EfficientNet models each trained using five-fold cross-validation. Data augmentation techniques were applied during training to reduce the risk of overfitting. The method was initially trained on the 2019 challenge dataset, followed by training on the combined 2019 and 2020 challenge datasets. However, unlike ADAE, this method did not use any metadata during the training process.

\subsubsection{3rd Place,~\texorpdfstring{\cite{kaggleSIIMISICMelanoma}}{[3]}}
This approach employed an ensemble of eight EfficientNet models, each trained using different combinations of input image resolutions (256, 384, 512, and 768). Training also incorporated data from prior years of the challenge (2017–2019), along with hair augmentation techniques and patient metadata. Unlike the other two methods, no cross-validation was performed on any of the models.

\subsubsection{Expected Risk Minimization (ERM)~\texorpdfstring{\citep{Vapnik1999}}{[4]}}
\textcolor{black}{In line with other benchmarking work~\citep{zong2023medfairbenchmarkingfairnessmedical}, we also include a standard ERM algorithm that serves as a baseline to compare performance. For the ERM baseline, we used a ResNet-18 architecture with cross-entropy loss, trained from scratch using the Adam optimizer. Hyperparameters were selected via Bayesian optimization, with the search space including learning rates in the range $[1 \times 10^{-5}, 1 \times 10^{-3}]$, weight decay values of $1 \times 10^{-4}$ and $1 \times 10^{-5}$, and batch sizes of 256, 512, and 1024. No data augmentation techniques were applied and no metadata was incorporated into the training process. We held out 10\% of the ISIC 2020 training data as a validation set, which was used for model selection and early stopping. The final configuration uses a learning rate of $1.475\times 10^{-4}$, selected after being trained for 30 epochs with a batch size of 256. Model Selection was done based on the model that minimized validation loss.}

\subsection{Metrics and Evaluation} 

\subsubsection{Choice of Calibration Metric}

\cite{Feng2023IsTM} reframe strong calibration as a Score-Based CUSUM test that indicates if there is \emph{at least one} subgroup that is miscalibrated in a dataset~\citep{Feng2023IsTM}.
Let $\tilde{p}: X \mapsto [0,1]$  be the risk prediction algorithm and $p_0: X \mapsto [0,1]$ be the true event rate over some domain $X \in \mathbb{R}^{d}$, where d is the number of features. For some pre-specified tolerance level $\delta$, define a poorly calibrated subgroup as $A_\delta = \{ x \in X: \lvert \tilde{p}(x)-p_{0}(x) \rvert > \delta \}$. An AI model is \textbf{strongly calibrated} if there is no individual from a dataset that is mis-calibrated. 

Suppose the dataset is composed of $n$ independent and identically distributed (i.i.d) observations with samples $x_i \in X$ and binary outcome $Y_i$ for $i = 1, ..., n$. Let $\hat{p}: X \mapsto [0,1]$ be the AI model being audited, and let $\hat{p}_\delta(x_i) = [\hat{p}(x_i) + \delta]_{[0,1]}$  be its prediction on a sample $x_i$. \cite{Feng2023IsTM}'s method involves partitioning the dataset into two groups of size $n_1$ and $n_2$ respectively. $n_1$ samples are used to train a group of $K$ models to predict the true event rate $p_0(X)$ of each sample (these models are called \textit{residual models} in the original paper). These models are trained on the metadata and predictions from $\hat{p}$ for each sample. The remaining data is used to calculate a potential test statistic using the product of the differences in the observed and predicted values of the residuals, represented as $(Y_i - \hat{p}_{\delta}(x_i))$ and $(\hat{p}_{0}(x_i) - \hat{p}_{\delta}(x_i))$ respectively.
Specifically, the chosen test statistic is given by 

\footnotesize
\begin{align}
    \hat T_{n,>}^{(split)} = \max_{k=1,\dots,K} \frac{1}{n_2} \sum_{i=n_1 + 1}^n (Y_i - \hat p_{\delta}(x_i)) \hat{g}_{\lambda,k}(x)\mathbbm{1} \{ \hat{g}_{\lambda,k}(x) > 0 \}
\end{align}
\normalsize

where $\hat{g}_{\lambda,k}(x) = (\hat{p}_{0}(x) - \hat{p}_{\delta}(x))$ are the predicted residuals from the $kth$ residual model. A model is said to be \emph{miscalibrated} if the largest of these $k$ test statistics exceeds a threshold. 

In addition to this CUSUM test, this method uses Variable Importance plots that are generated by measuring how much the test statistic changes when the features that the residual model was trained on are modified. The most important features are the ones that greatly change the final result of the test when they are modified. Through this two-part process of first using the CUSUM test to check if there is \textit{at least} one miscalibrated group, then using the Variable Importance plots to see which groups they may be, this method is able to circumvent the issues of subgroup imbalance and multi-group membership that are often seen in calibration problems.



In order to improve the estimate of the true event rate, we provide intermediate feature embeddings from each $\hat{p}$ being audited as part of the training data to these residual models. For example, when auditing ADAE on the ISIC 2020 dataset, each sample used to train the residual models contained the original metadata as well as intermediate feature embeddings that were extracted while ADAE was making predictions on the ISIC 2020 test set. 

\begin{table*}[t]
\centering
\caption{Performance of the 4 models on the ISIC 2020 dataset, stratified by various demographic factors. The top 3 models outperform ERM overall. ADAE has a lower specificity than the other two models which may suggest that it tends to over-diagnose patients at a higher rate.}
\Large
\resizebox{\linewidth}{!}{
\begin{tabular}{|lrrcccccccccccccccc|}
\hline
\multicolumn{1}{|l|}{Characteristic}                    & \multicolumn{1}{l|}{Total lesions} & \multicolumn{1}{l|}{Total Melanomas} & \multicolumn{4}{l|}{False Positives at 95\% Threshold}                                                                     & \multicolumn{4}{l|}{Sensitivity at 95\% Threshold}                                                                                        & \multicolumn{4}{l|}{Specificity 95\% Threshold}                                                                                        & \multicolumn{4}{l|}{AUROC}                                                                                                \\ \hline
\multicolumn{1}{|l|}{}      & \multicolumn{2}{l|}{}                            & \multicolumn{1}{l|}{ADAE} & \multicolumn{1}{l|}{2nd Place} & \multicolumn{1}{l|}{3rd Place} & \multicolumn{1}{l|}{ERM}  & \multicolumn{1}{l|}{ADAE} & \multicolumn{1}{l|}{2nd Place} & \multicolumn{1}{l|}{3rd Place} & \multicolumn{1}{l|}{ERM}  & \multicolumn{1}{l|}{ADAE} & \multicolumn{1}{l|}{2nd Place} & \multicolumn{1}{l|}{3rd Place} & \multicolumn{1}{l|}{ERM}  & \multicolumn{1}{l|}{ADAE}  & \multicolumn{1}{l|}{2nd Place} & \multicolumn{1}{l|}{3rd Place} & \multicolumn{1}{l|}{ERM} \\ \hline

\multicolumn{1}{|c|}{Overall}            & \multicolumn{1}{c|}{10982}          & \multicolumn{1}{c|}{261}              & \multicolumn{1}{c|}{3157} & \multicolumn{1}{c|}{2469}      & \multicolumn{1}{c|}{2294}      & \multicolumn{1}{c|}{5864} & \multicolumn{1}{c|}{95\%} & \multicolumn{1}{c|}{95\%}      & \multicolumn{1}{c|}{95\%}      & \multicolumn{1}{c|}{95\%} & \multicolumn{1}{c|}{71\%} & \multicolumn{1}{c|}{77\%}      & \multicolumn{1}{c|}{79\%}      & \multicolumn{1}{c|}{45\%} & \multicolumn{1}{c|}{0.949} & \multicolumn{1}{c|}{0.949}     & \multicolumn{1}{c|}{0.953}     & 0.866                    \\ \hline

\multicolumn{19}{|l|}{Age}                                                                                                                                                                                                                                                                                                                                                                                                                                                                                                                                                                                                    \\ \hline
\multicolumn{1}{|c|}{Under 50}            & \multicolumn{1}{c|}{5306}          & \multicolumn{1}{c|}{67}              & \multicolumn{1}{c|}{1601} & \multicolumn{1}{c|}{1113}      & \multicolumn{1}{c|}{1427}      & \multicolumn{1}{c|}{3002} & \multicolumn{1}{c|}{94\%} & \multicolumn{1}{c|}{94\%}      & \multicolumn{1}{c|}{95\%}      & \multicolumn{1}{c|}{97\%} & \multicolumn{1}{c|}{69\%} & \multicolumn{1}{c|}{79\%}      & \multicolumn{1}{c|}{77\%}      & \multicolumn{1}{c|}{43\%} & \multicolumn{1}{c|}{0.942} & \multicolumn{1}{c|}{0.933}     & \multicolumn{1}{c|}{0.949}     & 0.862                    \\ \hline
\multicolumn{1}{|c|}{Over or Equal to 50} & \multicolumn{1}{c|}{5676}          & \multicolumn{1}{c|}{194}             & \multicolumn{1}{c|}{1556} & \multicolumn{1}{c|}{1356}      & \multicolumn{1}{c|}{867}       & \multicolumn{1}{c|}{2862} & \multicolumn{1}{c|}{94\%} & \multicolumn{1}{c|}{95\%}      & \multicolumn{1}{c|}{90\%}      & \multicolumn{1}{c|}{94\%} & \multicolumn{1}{c|}{69\%} & \multicolumn{1}{c|}{75\%}      & \multicolumn{1}{c|}{83\%}      & \multicolumn{1}{c|}{48\%} & \multicolumn{1}{c|}{0.951} & \multicolumn{1}{c|}{0.952}     & \multicolumn{1}{c|}{0.949}     & 0.867                    \\ \hline
\multicolumn{19}{|l|}{Sex}                                                                                                                                                                                                                                                                                                                                                                                                                                                                                                                                                                                                    \\ \hline
\multicolumn{1}{|c|}{Female}              & \multicolumn{1}{c|}{4727}          & \multicolumn{1}{c|}{90}              & \multicolumn{1}{c|}{1307} & \multicolumn{1}{c|}{1018}      & \multicolumn{1}{c|}{867}       & \multicolumn{1}{c|}{2452} & \multicolumn{1}{c|}{93\%} & \multicolumn{1}{c|}{92\%}      & \multicolumn{1}{c|}{91\%}      & \multicolumn{1}{c|}{94\%} & \multicolumn{1}{c|}{72\%} & \multicolumn{1}{c|}{78\%}      & \multicolumn{1}{c|}{81\%}      & \multicolumn{1}{c|}{47\%} & \multicolumn{1}{c|}{0.946} & \multicolumn{1}{c|}{0.940}     & \multicolumn{1}{c|}{0.952}     & 0.872                    \\ \hline
\multicolumn{1}{|c|}{Male}                & \multicolumn{1}{c|}{6255}          & \multicolumn{1}{c|}{171}             & \multicolumn{1}{c|}{1850} & \multicolumn{1}{c|}{1451}      & \multicolumn{1}{c|}{1427}      & \multicolumn{1}{c|}{3412} & \multicolumn{1}{c|}{96\%} & \multicolumn{1}{c|}{96\%}      & \multicolumn{1}{c|}{97\%}      & \multicolumn{1}{c|}{95\%} & \multicolumn{1}{c|}{70\%} & \multicolumn{1}{c|}{76\%}      & \multicolumn{1}{c|}{77\%}      & \multicolumn{1}{c|}{44\%} & \multicolumn{1}{c|}{0.951} & \multicolumn{1}{c|}{0.954}     & \multicolumn{1}{c|}{0.953}     & 0.863                    \\ \hline
\end{tabular}}

\label{Tab:95_pct_ISIC}
\end{table*}


\begin{table*}[t]
\centering
\caption{Difference in AUROC for each model on the ISIC 2020 dataset using DeLong's correlated test. Significant (p $<$ 0.05, blue) and marginally significant (0.05 $\leq$ p $\leq$ 0.1, yellow) results are highlighted. The top 3 models all had similar AUROCs on ISIC 2020, and all were much significantly better than ERM.}
\begin{subtable}{\textwidth}{
\centering
\resizebox{\linewidth}{!}{
\begin{tabular}{|c|cc|cc|cc|}
    \hline
    \multicolumn{1}{|l|}{} & \multicolumn{2}{c|}{\textbf{ADAE vs 2nd Place}}                              & \multicolumn{2}{c|}{\textbf{2nd Place vs 3rd Place}}                                   & \multicolumn{2}{c|}{\textbf{ADAE vs 3rd Place}}                              \\ \hline
    \textbf{Subgroup}      & \multicolumn{1}{c|}{\textbf{Difference in AUROC (95\% CI)}} & \textbf{P-value} & \multicolumn{1}{c|}{\textbf{Difference in AUROC (95\% CI)}}           & \textbf{P-value} & \multicolumn{1}{c|}{\textbf{Difference in AUROC (95\% CI)}} & \textbf{P-value} \\ \hline
    Everyone               & \multicolumn{1}{c|}{0.001 (-0.007, 0.008)}                & 0.856            & \multicolumn{1}{c|}{-0.004 (-0.011  0.004)} & 0.333            & \multicolumn{1}{c|}{-0.003 (-0.011, 0.004)}                & 0.421            \\ \hline
    Men                    & \multicolumn{1}{c|}{-0.003 (-0.011, 0.005)}               & 0.493            & \multicolumn{1}{c|}{0.001 (-0.009, 0.011)}                          & 0.856            & \multicolumn{1}{c|}{-0.002 (-0.011, 0.007)}               & 0.694            \\ \hline
    Women                  & \multicolumn{1}{c|}{0.006 (-0.008, 0.020)}                & 0.435            & \multicolumn{1}{c|}{-0.012 (-0.024, 0.001)}                         & 0.066            & \multicolumn{1}{c|}{-0.001 (-0.019,  0.007)}              & 0.379            \\ \hline
    Men, Age Over 50       & \multicolumn{1}{c|}{-0.002 (-0.009, 0.006)}               & 0.518            & \multicolumn{1}{c|}{\cellcolor{yellow!25}0.011 (-0.001,  0.022)}                         & \cellcolor{yellow!25}0.081            & \multicolumn{1}{c|}{\cellcolor{yellow!25}0.001 (-0.001, 0.019)}                & \cellcolor{yellow!25}0.092            \\ \hline
    Women, Age Over 50     & \multicolumn{1}{c|}{0.000 (-0.018, 0.018)}                & 0.980            & \multicolumn{1}{c|}{-0.012 (-0.026, 0.003)}                         & 0.115            & \multicolumn{1}{c|}{-0.011 (-0.027, 0.004)}               & 0.159            \\ \hline
    Men, Age Under 50      & \multicolumn{1}{c|}{-0.001 (-0.026, 0.024)}               & 0.945            & \multicolumn{1}{c|}{-0.023 (-0.052, 0.006)}                         & 0.122            & \multicolumn{1}{c|}{-0.024 (-0.053, 0.006)}               & 0.114            \\ \hline
    Women, Age Under 50    & \multicolumn{1}{c|}{\cellcolor{yellow!25}0.018 (-0.002, 0.038)}                & \cellcolor{yellow!25}0.075            & \multicolumn{1}{c|}{-0.009 (-0.026, 0.008)}                         & 0.303            & \multicolumn{1}{c|}{0.009 (-0.007, 0.025)}                & 0.257            \\ \hline
\end{tabular}}}

\label{Tab:Diff_AUROC_model_ISIC_a}
\end{subtable}
\hspace{0.5\textwidth}

\begin{subtable}{\textwidth}{
\centering
\large
\resizebox{\linewidth}{!}{
\begin{tabular}{|c|cc|cc|cc|}
\hline
\multicolumn{1}{|l|}{} & \multicolumn{2}{c|}{\textbf{ADAE vs ERM}}                              & \multicolumn{2}{c|}{\textbf{2nd Place vs ERM}}                                   & \multicolumn{2}{c|}{\textbf{3rd Place vs ERM}}                              \\ \hline
\textbf{Subgroup}      & \multicolumn{1}{c|}{\textbf{Difference in AUROC (95\% CI)}} & \textbf{P-value} & \multicolumn{1}{c|}{\textbf{Difference in AUROC (95\% CI)}}           & \textbf{P-value} & \multicolumn{1}{c|}{\textbf{Difference in AUROC (95\% CI)}} & \textbf{P-value} \\ \hline
Everyone               & \multicolumn{1}{c|}{\cellcolor{blue!25}0.084 (0.065, 0.102)}                & \cellcolor{blue!25}0.000            & \multicolumn{1}{c|}{\cellcolor{blue!25}0.083 (0.065, 0.101)} & \cellcolor{blue!25}0.000            & \multicolumn{1}{c|}{\cellcolor{blue!25}0.087 (0.068, 0.105)}                & \cellcolor{blue!25}0.000            \\ \hline
Men                    & \multicolumn{1}{c|}{\cellcolor{blue!25}0.088 (0.066 , 0.111)}               & \cellcolor{blue!25}0.000            & \multicolumn{1}{c|}{\cellcolor{blue!25}0.091 (0.07, 0.113)}                          & \cellcolor{blue!25}0.000            & \multicolumn{1}{c|}{\cellcolor{blue!25}0.09 (0.068, 0.112)}               & \cellcolor{blue!25}0.000            \\ \hline
Women                  & \multicolumn{1}{c|}{\cellcolor{blue!25}0.074 (0.042 , 0.106)}                & \cellcolor{blue!25}0.000            & \multicolumn{1}{c|}{\cellcolor{blue!25}0.068 (0.037, 0.1)}                         & \cellcolor{blue!25}0.000            & \multicolumn{1}{c|}{\cellcolor{blue!25}0.08 (0.048, 0.112)}              & \cellcolor{blue!25}0.000            \\ \hline
Men, Age Over 50       & \multicolumn{1}{c|}{\cellcolor{blue!25}0.09 (0.065, 0.114)}               & \cellcolor{blue!25}0.000            & \multicolumn{1}{c|}{\cellcolor{blue!25}0.091 (0.069, 0.114)}                         & \cellcolor{blue!25}0.000            & \multicolumn{1}{c|}{\cellcolor{blue!25}0.081 (0.057, 0.104)}                & \cellcolor{blue!25}0.000            \\ \hline
Women, Age Over 50     & \multicolumn{1}{c|}{\cellcolor{blue!25}0.075 (0.031, 0.118)}                & \cellcolor{blue!25}0.001            & \multicolumn{1}{c|}{\cellcolor{blue!25}0.075 (0.031, 0.118)}                         & \cellcolor{blue!25}0.001            & \multicolumn{1}{c|}{\cellcolor{blue!25}0.086 (0.045, 0.128)}               & \cellcolor{blue!25}0.000            \\ \hline
Men, Age Under 50      & \multicolumn{1}{c|}{\cellcolor{blue!25}0.086 (0.035, 0.136)}               & \cellcolor{blue!25}0.001            & \multicolumn{1}{c|}{\cellcolor{blue!25}0.087 (0.033, 0.14)}                         & \cellcolor{blue!25}0.002            & \multicolumn{1}{c|}{\cellcolor{blue!25}0.109 (0.058, 0.16)}               & \cellcolor{blue!25}0.000            \\ \hline
Women, Age Under 50    & \multicolumn{1}{c|}{\cellcolor{blue!25}0.072 (0.03, 0.114)}                & \cellcolor{blue!25}0.001            & \multicolumn{1}{c|}{\cellcolor{blue!25}0.054 (0.013, 0.095)}                         & \cellcolor{blue!25}0.010            & \multicolumn{1}{c|}{\cellcolor{blue!25}0.063 (0.016, 0.11)}                & \cellcolor{blue!25}0.009            \\ \hline
\end{tabular}}}
\label{Tab:Diff_AUROC_model_ISIC_b}
\end{subtable}

\label{Tab:Diff_AUROC_model_ISIC}
\end{table*}

\subsubsection{Statistical Analysis}
Model discrimination was summarized with AUROC, which is a commonly used metric for clinical binary classification. We evaluate the participating models from two aspects of AUC: 
\begin{enumerate}
    \item \textbf{Utility}: AUC gap across different models on the same subgroup (i.e. comparing the AUC of ADAE to that of 2nd Place for women); and
    \item \textbf{Group Fairness}: AUC gap between different subgroups that were evaluated by the same model (i.e. comparing the AUC of women under 50 to that of men under 50 for ADAE).
\end{enumerate}

For each form of analysis, we use Delong's test~\citep{DeLong1988ComparingTA}. DeLong's test is a statistical method used to compare the AUROC curves of two or more models. This test helps determine if there is a significant difference between the AUROC values of the models, which can indicate whether one model performs better than another in terms of classification accuracy and discriminative ability. The test is based on the Mann-Whitney U statistic~\citep{Mann1947}, which is equivalent to the empirical AUROC~\citep{Bitterlich2003}. It involves calculating the variance of the AUROC and using it to assess the statistical significance of the difference between the AUROC values of the models.

\textcolor{black}{DeLong’s test for uncorrelated ROC curves was used for utility analyses, and DeLong’s test for correlated ROC curves was used for group fairness analyses. The correlated version of DeLong's test for AUROC is used when the same data is used to generate both AUROC curves. It assumes that the predictions from the two models are correlated because they are based on the same data. 
On the other hand, the uncorrelated version of DeLong's test is used when different sets of data are used to generate the AUROC curves. It assumes that the predictions from the two models are independent because they are based on different sets of data.} 

The chosen level of significance was 0.05, and analyses were performed in R~\citep{Rsoftware}.

\section{Results} \label{sec: Results}


In this section, we first present our results on each dataset for Model Discrimination, followed by Model Calibration (Sections~\ref{subsec: MD} and~\ref{subsec: MC}, respectively).

\subsection{Model Discrimination} \label{subsec: MD}
\subsubsection{ISIC 2020}





\textcolor{black}{Table~\ref{Tab:95_pct_ISIC} summarizes the performance of the four evaluated models at a 95\% sensitivity threshold. While all three ISIC 2020 challenge models outperformed the ERM baseline in terms of AUROC, ADAE exhibited the highest number of false positives across subgroups, suggesting a tendency to over-diagnose.}

\textcolor{black}{Tables~\ref{Tab:Diff_AUROC_model_ISIC} and~\ref{Tab:Diff_AUROC_subgroup_ISIC} present AUROC comparisons across demographic subgroups. Differences in model discrimination between the top three models were generally small and not statistically significant (all $p$-values $>$ 0.05), with only a few subgroup comparisons showing marginal differences. In contrast, all three models significantly outperformed ERM across every subgroup, reinforcing their overall superiority in discrimination performance.}

\textcolor{black}{These results indicate that while the top-performing models are comparable in overall accuracy, their subgroup-specific behavior—particularly ADAE’s lower specificity—warrants further scrutiny in clinical contexts.}

\subsubsection{PROVE-AI}


\begin{table}[t]
    \centering
    \caption{Difference in AUROC for each subgroup on the ISIC 2020 dataset using DeLong's uncorrelated Test. Each comparison produced a difference that was not significant.}
    \resizebox{\columnwidth}{!}{
    \begin{tabular}{|cl|l|c|c|c|}
        \hline
        \multicolumn{2}{|c|}{\textbf{Model}} & \multicolumn{1}{c|}{\textbf{Subgroup}} & \multicolumn{1}{c|}{\textbf{Women AUROC}} & \multicolumn{1}{c|}{\textbf{Men AUROC}} & \multicolumn{1}{c|}{\textbf{P-value}} \\ \hline
        \multicolumn{2}{|c|}{ADAE}           & Age Under 50                           & 0.955                                   & 0.931                                 & 0.439                                 \\ \hline
        \multicolumn{2}{|c|}{2nd Place}      & Age Under 50                           & 0.937                                   & 0.932                                 & 0.892                                 \\ \hline
        \multicolumn{2}{|c|}{3rd Place}      & Age Under 50                           & 0.945                                   & 0.955                                 & 0.665                                 \\ \hline
        \multicolumn{2}{|c|}{ERM}            & Age Under 50                           & 0.882                                   & 0.846                                 & 0.385                                 \\ \hline
        \multicolumn{2}{|c|}{ADAE}           & Age Over 50    & 0.942           & 0.955         & 0.486                                 \\ \hline
        \multicolumn{2}{|c|}{2nd Place}      & Age Over 50    & 0.941                                   & 0.956                                 & 0.465                                 \\ \hline
        \multicolumn{2}{|c|}{3rd Place}      & Age Over 50    & 0.953                                   & 0.946                                 & 0.646                                 \\ \hline
        \multicolumn{2}{|c|}{ERM}            & Age Over 50    & 0.867                                   & 0.865                                 & 0.955                                 \\ \hline
    \end{tabular}}
    \label{Tab:Diff_AUROC_subgroup_ISIC}
\end{table}

\begin{table}[t]
\centering
\caption{Difference in AUROC for each subgroup on the PROVE-AI dataset using DeLong's uncorrelated Test.  One marginally significant (0.05 $\leq$ p $\leq$ 0.1, yellow) result was produced from this test, with every other comparison producing a difference that was not significant.}
\resizebox{\columnwidth}{!}{
\begin{tabular}{|cl|l|c|c|c|}
\hline
\multicolumn{2}{|c|}{\textbf{Model}} & \multicolumn{1}{c|}{\textbf{Subgroup}} & \multicolumn{1}{c|}{\textbf{FST I AUROC}} & \multicolumn{1}{c|}{\textbf{FST II AUROC}} & \multicolumn{1}{c|}{\textbf{P-value}} \\ \hline
\multicolumn{2}{|c|}{ADAE}           & Men                                    & 0.89                                    & 0.85                                     & 0.67                                  \\ \hline
\multicolumn{2}{|c|}{2nd Place}      & \cellcolor{yellow!25}Men                                    & \cellcolor{yellow!25}0.93                                    & \cellcolor{yellow!25}0.81                                     & \cellcolor{yellow!25}0.09                                  \\ \hline
\multicolumn{2}{|c|}{3rd Place}      & Men                                    & 0.62                                    & 0.70                                     & 0.66                                  \\ \hline
\multicolumn{2}{|c|}{ERM}            & Men                                    & 0.52                                    & 0.70                                     & 0.27                                  \\ \hline
\multicolumn{2}{|c|}{ADAE}           & Women          & 0.82                                    & 0.77                                     & 0.76                                  \\ \hline
\multicolumn{2}{|c|}{2nd Place}      & Women          & 0.77                                    & 0.72                                     & 0.64                                  \\ \hline
\multicolumn{2}{|c|}{3rd Place}      & Women          & 0.58                                    & 0.64                                     & 0.74                                  \\ \hline
\multicolumn{2}{|c|}{ERM}            & Women          & 0.67                                    & 0.74                                     & 0.58                                  \\ \hline
\multicolumn{2}{|c|}{ADAE}           & Age Under 65                           & 0.80                                    & 0.72                                     & 0.58                                  \\ \hline
\multicolumn{2}{|c|}{2nd Place}      & Age Under 65                           & 0.75                                    & 0.65                                     & 0.72                                  \\ \hline
\multicolumn{2}{|c|}{3rd Place}      & Age Under 65                           & 0.64                                    & 0.47                                     & 0.50                                  \\ \hline
\multicolumn{2}{|c|}{ERM}            & Age Under 65                           & 0.73                                    & 0.48                                     & 0.31                                  \\ \hline
\multicolumn{2}{|c|}{ADAE}           & Age Over 65    & 0.86                                    & 0.83                                     & 0.76                                  \\ \hline
\multicolumn{2}{|c|}{2nd Place}      & Age Over 65    & 0.83                                    & 0.80                                     & 0.84                                  \\ \hline
\multicolumn{2}{|c|}{3rd Place}      & Age Over 65    & 0.71                                    & 0.77                                     & 0.64                                  \\ \hline
\multicolumn{2}{|c|}{ERM}            & Age Over 65    & 0.72                                    & 0.73                                     & 0.94                                  \\ \hline
\end{tabular}}

\label{Tab:Diff_AUROC_subgroup_Prove}
\end{table}

Tables~\ref{Tab:95_pct_Prove},~\ref{Tab:Diff_AUROC_model_Prove}, and~\ref{Tab:Diff_AUROC_subgroup_Prove} show the same respective analysis as Tables~\ref{Tab:95_pct_ISIC},~\ref{Tab:Diff_AUROC_model_ISIC}, and~\ref{Tab:Diff_AUROC_subgroup_ISIC}, but now applied to the PROVE-AI dataset which unlike ISIC 2020 features information about the Fitzpatrick Skin Tone (FST) of each patient. 



The top 3 once again outperformed ERM, and on this dataset we see ADAE has the best performance of the top 3 models in terms of False Positives, Specificity and AUROC. At this 95\% threshold ADAE had a higher specificity than 2nd Place (40\% vs 26\%) and 3rd Place (40\% vs 10\%). 3rd Place was the model that had the most False Positives for every subgroup. This may suggest that of these 4 models, ADAE is best equipped to perform on patients that are of a different distribution than the one it was trained on.

\begin{table*}[t]
\centering
\caption{Performance of the 4 models on the PROVE-AI dataset, stratified by various demographic factors. The performance of each model is generally worse than it was on ISIC 2020. In particular, the performance of the 3rd Place model is close to ERM, which is further exemplified in the other tables of this section.}
\large
\resizebox{\linewidth}{!}{
\begin{tabular}{|ccccccccccccccccccc|}
\hline
\multicolumn{1}{|l|}{Characteristic}                    & \multicolumn{1}{l|}{Total lesions} & \multicolumn{1}{l|}{Total Melanomas} & \multicolumn{4}{l|}{False Positives at 95\% Threshold}                                                                     & \multicolumn{4}{l|}{Sensitivity at 95\% Threshold}                                                                                        & \multicolumn{4}{l|}{Specificity at 95\% Threshold}                                                                                        & \multicolumn{4}{l|}{AUROC}                                                                                                \\ \hline
\multicolumn{1}{|l|}{}      & \multicolumn{2}{l|}{}                            & \multicolumn{1}{l|}{ADAE} & \multicolumn{1}{l|}{2nd Place} & \multicolumn{1}{l|}{3rd Place} & \multicolumn{1}{l|}{ERM}  & \multicolumn{1}{l|}{ADAE} & \multicolumn{1}{l|}{2nd Place} & \multicolumn{1}{l|}{3rd Place} & \multicolumn{1}{l|}{ERM}  & \multicolumn{1}{l|}{ADAE} & \multicolumn{1}{l|}{2nd Place} & \multicolumn{1}{l|}{3rd Place} & \multicolumn{1}{l|}{ERM}  & \multicolumn{1}{l|}{ADAE}  & \multicolumn{1}{l|}{2nd Place} & \multicolumn{1}{l|}{3rd Place} & \multicolumn{1}{l|}{ERM} \\ \hline

\multicolumn{1}{|c|}{Overall}            & \multicolumn{1}{c|}{603}          & \multicolumn{1}{c|}{95}              & \multicolumn{1}{c|}{303} & \multicolumn{1}{c|}{375}      & \multicolumn{1}{c|}{457}      & \multicolumn{1}{c|}{445} & \multicolumn{1}{c|}{96\%} & \multicolumn{1}{c|}{96\%}      & \multicolumn{1}{c|}{96\%}      & \multicolumn{1}{c|}{96\%} & \multicolumn{1}{c|}{40\%} & \multicolumn{1}{c|}{26\%}      & \multicolumn{1}{c|}{10\%}      & \multicolumn{1}{c|}{12\%} & \multicolumn{1}{c|}{0.850} & \multicolumn{1}{c|}{0.819}     & \multicolumn{1}{c|}{0.704}     & 0.735                    \\ \hline

\multicolumn{19}{|l|}{Age}                                                                                                                                                                                                                                                                                                                                                                                                                                                                                                                                                                                                                                                                                                                                                                                                                                                                                                                                                                         \\ \hline
\multicolumn{1}{|c|}{Under 65}            & \multicolumn{1}{c|}{307}                                                & \multicolumn{1}{c|}{32}                                                & \multicolumn{1}{c|}{127}                        & \multicolumn{1}{c|}{177}                        & \multicolumn{1}{c|}{248}                        & \multicolumn{1}{c|}{233}                        & \multicolumn{1}{c|}{94\%}                         & \multicolumn{1}{c|}{91\%}                                                 & \multicolumn{1}{c|}{94\%}      & \multicolumn{1}{c|}{97\%}  & \multicolumn{1}{c|}{54\%} & \multicolumn{1}{c|}{36\%}      & \multicolumn{1}{c|}{10\%}      & \multicolumn{1}{c|}{15\%} & \multicolumn{1}{c|}{0.853} & \multicolumn{1}{c|}{0.828}                         & \multicolumn{1}{c|}{0.702}     & 0.676                    \\ \hline
\multicolumn{1}{|c|}{Over or Equal to 65} & \multicolumn{1}{c|}{296}                                                & \multicolumn{1}{c|}{63}                                                & \multicolumn{1}{c|}{176}                                                & \multicolumn{1}{c|}{198}                                                & \multicolumn{1}{c|}{209}                                                & \multicolumn{1}{c|}{212}                                                & \multicolumn{1}{c|}{97\%}                                                 & \multicolumn{1}{c|}{98\%}                                                 & \multicolumn{1}{c|}{97\%}      & \multicolumn{1}{c|}{95\%}  & \multicolumn{1}{c|}{24\%} & \multicolumn{1}{c|}{15\%}      & \multicolumn{1}{c|}{10\%}      & \multicolumn{1}{c|}{9\%}  & \multicolumn{1}{c|}{0.833} & \multicolumn{1}{c|}{0.784} & \multicolumn{1}{c|}{0.684}     & 0.743                    \\ \hline
\multicolumn{19}{|l|}{Sex}                                                                                                                                                                                                                                                                                                                                                                                                                                                                                                                                                                                                                                                                                                                                                                                                                                                                                                                                                                         \\ \hline
\multicolumn{1}{|c|}{Female}              & \multicolumn{1}{c|}{{\color[HTML]{222222} 324}} & \multicolumn{1}{c|}{{\color[HTML]{222222} 32}} & \multicolumn{1}{c|}{163}                                                & \multicolumn{1}{c|}{170}                                                & \multicolumn{1}{c|}{201}                                                & \multicolumn{1}{c|}{179}                                                & \multicolumn{1}{c|}{94\%}                                                 & \multicolumn{1}{c|}{97\%}                                                 & \multicolumn{1}{c|}{97\%}      & \multicolumn{1}{c|}{94\%}  & \multicolumn{1}{c|}{44\%} & \multicolumn{1}{c|}{30\%}      & \multicolumn{1}{c|}{12\%}      & \multicolumn{1}{c|}{9\%}  & \multicolumn{1}{c|}{0.804} & \multicolumn{1}{c|}{0.770}                         & \multicolumn{1}{c|}{0.647}     & 0.720                    \\ \hline
\multicolumn{1}{|c|}{Male}                & \multicolumn{1}{c|}{{\color[HTML]{222222} 279}} & \multicolumn{1}{c|}{{\color[HTML]{222222} 63}} & \multicolumn{1}{c|}{140}                                                & \multicolumn{1}{c|}{205}                                                & \multicolumn{1}{c|}{256}                                                & \multicolumn{1}{c|}{266}                                                & \multicolumn{1}{c|}{97\%}                                                 & \multicolumn{1}{c|}{95\%}                                                 & \multicolumn{1}{c|}{95\%}      & \multicolumn{1}{c|}{97\%}  & \multicolumn{1}{c|}{35\%} & \multicolumn{1}{c|}{21\%}      & \multicolumn{1}{c|}{7\%}       & \multicolumn{1}{c|}{17\%} & \multicolumn{1}{c|}{0.865} & \multicolumn{1}{c|}{0.835}                         & \multicolumn{1}{c|}{0.720}     & 0.735                    \\ \hline
\multicolumn{19}{|l|}{FST}                                                                                                                                                                                                                                                                                                                                                                                                                                                                                                                                                                                                                                                                                                                                                                                                                                                                                                                                                                         \\ \hline
\multicolumn{1}{|c|}{I}                   & \multicolumn{1}{c|}{{\color[HTML]{222222} 46}}  & \multicolumn{1}{c|}{{\color[HTML]{222222} 16}} & \multicolumn{1}{c|}{{\color[HTML]{222222} 19}}  & \multicolumn{1}{c|}{{\color[HTML]{222222} 27}}  & \multicolumn{1}{c|}{{\color[HTML]{222222} 33}}  & \multicolumn{1}{c|}{{\color[HTML]{222222} 35}}  & \multicolumn{1}{c|}{{\color[HTML]{222222} 100\%}} & \multicolumn{1}{c|}{{\color[HTML]{222222} 100\%}} & \multicolumn{1}{c|}{89\%}      & \multicolumn{1}{c|}{89\%}  & \multicolumn{1}{c|}{49\%} & \multicolumn{1}{c|}{27\%}      & \multicolumn{1}{c|}{11\%}      & \multicolumn{1}{c|}{5\%}  & \multicolumn{1}{c|}{0.856} & \multicolumn{1}{c|}{0.874}                         & \multicolumn{1}{c|}{0.628}     & 0.568                    \\ \hline
\multicolumn{1}{|c|}{II}                  & \multicolumn{1}{c|}{{\color[HTML]{222222} 333}} & \multicolumn{1}{c|}{{\color[HTML]{222222} 57}} & \multicolumn{1}{c|}{{\color[HTML]{222222} 171}} & \multicolumn{1}{c|}{{\color[HTML]{222222} 214}} & \multicolumn{1}{c|}{{\color[HTML]{222222} 253}} & \multicolumn{1}{c|}{{\color[HTML]{222222} 243}} & \multicolumn{1}{c|}{{\color[HTML]{222222} 94\%}}  & \multicolumn{1}{c|}{{\color[HTML]{222222} 92\%}}  & \multicolumn{1}{c|}{96\%}      & \multicolumn{1}{c|}{98\%}  & \multicolumn{1}{c|}{39\%} & \multicolumn{1}{c|}{24\%}      & \multicolumn{1}{c|}{10\%}      & \multicolumn{1}{c|}{14\%} & \multicolumn{1}{c|}{0.831} & \multicolumn{1}{c|}{0.790}                         & \multicolumn{1}{c|}{0.694}     & 0.723                    \\ \hline
\multicolumn{1}{|c|}{III}                 & \multicolumn{1}{c|}{{\color[HTML]{222222} 202}} & \multicolumn{1}{c|}{{\color[HTML]{222222} 19}} & \multicolumn{1}{c|}{{\color[HTML]{222222} 101}} & \multicolumn{1}{c|}{{\color[HTML]{222222} 123}} & \multicolumn{1}{c|}{{\color[HTML]{222222} 154}} & \multicolumn{1}{c|}{{\color[HTML]{222222} 151}} & \multicolumn{1}{c|}{{\color[HTML]{222222} 97\%}}  & \multicolumn{1}{c|}{{\color[HTML]{222222} 100\%}} & \multicolumn{1}{c|}{100\%}     & \multicolumn{1}{c|}{94\%}  & \multicolumn{1}{c|}{41\%} & \multicolumn{1}{c|}{28\%}      & \multicolumn{1}{c|}{10\%}      & \multicolumn{1}{c|}{12\%} & \multicolumn{1}{c|}{0.871} & \multicolumn{1}{c|}{0.835}                         & \multicolumn{1}{c|}{0.739}     & 0.800                    \\ \hline
\multicolumn{1}{|c|}{IV}                  & \multicolumn{1}{c|}{{\color[HTML]{222222} 22}}  & \multicolumn{1}{c|}{{\color[HTML]{222222} 3}}  & \multicolumn{1}{c|}{12}                                                 & \multicolumn{1}{c|}{11}                                                 & \multicolumn{1}{c|}{17}                                                 & \multicolumn{1}{c|}{{\color[HTML]{222222} 16}}  & \multicolumn{1}{c|}{100\%}                                                & \multicolumn{1}{c|}{100\%}                                                & \multicolumn{1}{c|}{67\%}      & \multicolumn{1}{c|}{100\%} & \multicolumn{1}{c|}{37\%} & \multicolumn{1}{c|}{42\%}      & \multicolumn{1}{c|}{11\%}      & \multicolumn{1}{c|}{16\%} & \multicolumn{1}{c|}{0.947} & \multicolumn{1}{c|}{0.947}                         & \multicolumn{1}{c|}{0.684}     & 0.807                    \\ \hline
\end{tabular}}

\label{Tab:95_pct_Prove}
\end{table*}

\begin{table*}[t]
\centering
\caption{Difference in AUROC for each model on the PROVE-AI dataset using DeLong's correlated Test.  Significant (p $<$ 0.05, blue) and marginally significant (0.05 $\leq$ p $\leq$ 0.1, yellow) results are highlighted. While ADAE has the best overall performance, it and the other top models sometimes do not perform significantly better than ERM.}
\begin{subtable}{\textwidth}{
\resizebox{\linewidth}{!}{
\begin{tabular}{|c|cc|cc|cc|}
\hline
\multicolumn{1}{|l|}{} & \multicolumn{2}{c|}{\textbf{ADAE vs 2nd Place}}                                                            & \multicolumn{2}{c|}{\textbf{2nd Place vs 3rd Place}}                                                        & \multicolumn{2}{c|}{\textbf{ADAE vs 3rd Place}}                                                             \\ \hline
\textbf{Subgroup}               & \multicolumn{1}{c|}{\textbf{Difference in AUROC (95\% CI)} }                 & \textbf{P-value}                       & \multicolumn{1}{c|}{\textbf{Difference in AUROC (95\% CI)}}                  & \textbf{P-value }                       & \multicolumn{1}{c|}{\textbf{Difference in AUROC (95\% CI)}}                   & \textbf{P-value}                       \\ \hline
Everyone               & \multicolumn{1}{c|}{\cellcolor{blue!25}0.031 (0.009, 0.053)} & \cellcolor{blue!25}0.006 & \multicolumn{1}{c|}{\cellcolor{blue!25}0.115 (0.055, 0.175)} & \cellcolor{blue!25}0.0001 & \multicolumn{1}{c|}{\cellcolor{blue!25}0.146 (0.087, 0.205)}                          & \cellcolor{blue!25}0.00                     \\ \hline
FST I Men              & \multicolumn{1}{c|}{-0.044 (-0.169, 0.080)}                       & 0.485                         & \multicolumn{1}{c|}{\cellcolor{yellow!25}0.311 (-0.021,  0.643)}                       & \cellcolor{yellow!25}0.066                          & \multicolumn{1}{c|}{0.267 (-0.061,  0.594)}                        & 0.110                         \\ \hline
FST II Men             & \multicolumn{1}{c|}{\cellcolor{blue!25}0.044 (0.004, 0.084)} & \cellcolor{blue!25}0.031 & \multicolumn{1}{c|}{\cellcolor{yellow!25}0.111 (-0.003,  0.225)}                       & \cellcolor{yellow!25}0.058                          & \multicolumn{1}{c|}{\cellcolor{blue!25}0.155 (0.054, 0.256)}  & \cellcolor{blue!25}0.003 \\ \hline
FST I Women            & \multicolumn{1}{c|}{0.045 (-0.169, 0.260)}                        & 0.678                         & \multicolumn{1}{c|}{0.197 (-0.065,  0.459)}                       & 0.140                          & \multicolumn{1}{c|}{\cellcolor{blue!25}0.242 ( 0.070, 0.415)} & \cellcolor{blue!25}0.006 \\ \hline
FST II Women           & \multicolumn{1}{c|}{\cellcolor{blue!25}0.051 (0.006, 0.095)} & \cellcolor{blue!25}0.026 & \multicolumn{1}{c|}{0.077 (-0.069,  0.223)}                       & 0.302                          & \multicolumn{1}{c|}{\cellcolor{yellow!25}0.128 (-0.0234,  0.279)}                       & \cellcolor{yellow!25}0.098                         \\ \hline
FST I, Age Under 65    & \multicolumn{1}{c|}{-0.035 (-0.163,  0.092)}                      & 0.589                         & \multicolumn{1}{c|}{0.247 (-0.049, 0.543)}                        & 0.102                          & \multicolumn{1}{c|}{0.212 (-0.081, 0.504)}                         & 0.156                         \\ \hline
FST II, Age Under 65   & \multicolumn{1}{c|}{\cellcolor{yellow!25}0.065 (-0.002, 0.132)}                        & \cellcolor{yellow!25}0.057                         & \multicolumn{1}{c|}{0.057 (-0.093, 0.208)}                        & 0.454                          & \multicolumn{1}{c|}{\cellcolor{blue!25}0.122 (0.004, 0.241)}  & \cellcolor{blue!25}0.043 \\ \hline
FST I, Age Over 65     & \multicolumn{1}{c|}{0.038 (-0.128, 0.203)}                        & 0.657                         & \multicolumn{1}{c|}{0.225 (-0.087, 0.537)}                        & 0.157                          & \multicolumn{1}{c|}{\cellcolor{yellow!25}0.263 (-0.035, 0.638)}                         & \cellcolor{yellow!25}0.084                         \\ \hline
FST II, Age Over 65    & \multicolumn{1}{c|}{\cellcolor{blue!25}0.049 ( 0.006, 0.093)}                        & \cellcolor{blue!25}0.027                         & \multicolumn{1}{c|}{0.096 (-0.022, 0.214)}                        & 0.110                          & \multicolumn{1}{c|}{\cellcolor{blue!25}0.146 (0.031, 0.260)}  & \cellcolor{blue!25}0.012 \\ \hline
\end{tabular}
}}
\end{subtable}

\begin{subtable}{\textwidth}{
\resizebox{\linewidth}{!}{
\begin{tabular}{|c|cc|cc|cc|}
\hline
\multicolumn{1}{|l|}{} & \multicolumn{2}{c|}{\textbf{ADAE vs ERM}}                                                            & \multicolumn{2}{c|}{\textbf{2nd Place vs ERM}}                                                        & \multicolumn{2}{c|}{\textbf{3rd Place vs ERM}}                                                             \\ \hline
\textbf{Subgroup}               & \multicolumn{1}{c|}{\textbf{Difference in AUROC (95\% CI)} }                 & \textbf{P-value}                       & \multicolumn{1}{c|}{\textbf{Difference in AUROC (95\% CI)}}                  & \textbf{P-value }                       & \multicolumn{1}{c|}{\textbf{Difference in AUROC (95\% CI)}}                   & \textbf{P-value}                       \\ \hline
Everyone               & \multicolumn{1}{c|}{\cellcolor{blue!25}0.115 (0.061, 0.169)} & \cellcolor{blue!25}0.000 & \multicolumn{1}{c|}{\cellcolor{blue!25}0.084 (0.033, 0.135)} & \cellcolor{blue!25}0.001 & \multicolumn{1}{c|}{-0.031 (-0.107, 0.045)}                          & 0.425                     \\ \hline
FST I Men              & \multicolumn{1}{c|}{\cellcolor{blue!25}0.367 (0.131, 0.603)}                       & \cellcolor{blue!25}0.002                         & \multicolumn{1}{c|}{\cellcolor{blue!25}0.411 (0.157, 0.666)}                       & \cellcolor{blue!25}0.002                          & \multicolumn{1}{c|}{0.1 (-0.298, 0.498)}                        & 0.623                         \\ \hline
FST II Men             & \multicolumn{1}{c|}{\cellcolor{blue!25}0.149 (0.059, 0.24)} & \cellcolor{blue!25}0.001 & \multicolumn{1}{c|}{\cellcolor{blue!25}0.105 (0.024, 0.186)}                       & \cellcolor{blue!25}0.011                          & \multicolumn{1}{c|}{-0.005 (-0.143, 0.133)}  & 0.938 \\ \hline
FST I Women            & \multicolumn{1}{c|}{0.152 (-0.219, 0.522)}                        & 0.423                         & \multicolumn{1}{c|}{0.106 (-0.112, 0.324)}                       & 0.341                          & \multicolumn{1}{c|}{-0.091 (-0.481, 0.299)} & 0.648 \\ \hline
FST II Women           & \multicolumn{1}{c|}{0.032 (-0.093, 0.157)} & 0.614 & \multicolumn{1}{c|}{-0.019 (-0.144, 0.107)}                       & 0.771                          & \multicolumn{1}{c|}{-0.096 (-0.221, 0.03)}                       & 0.135                         \\ \hline
FST I, Age Under 65    & \multicolumn{1}{c|}{0.2 (-0.112, 0.512)}                      & 0.209                         & \multicolumn{1}{c|}{0.235 (-0.059, 0.53)}                        & 0.117                          & \multicolumn{1}{c|}{-0.012 (-0.491, 0.468)}                         & 0.962                         \\ \hline
FST II, Age Under 65   & \multicolumn{1}{c|}{\cellcolor{blue!25}0.233 (0.102, 0.364)}                        & \cellcolor{blue!25}0.000                         & \multicolumn{1}{c|}{\cellcolor{blue!25}0.168 (0.043, 0.294)}                        & \cellcolor{blue!25}0.009                          & \multicolumn{1}{c|}{0.111 (-0.072, 0.293)}  & 0.234 \\ \hline
FST I, Age Over 65     & \multicolumn{1}{c|}{\cellcolor{blue!25}0.425 (0.195, 0.655)}                        & \cellcolor{blue!25}0.000                         & \multicolumn{1}{c|}{\cellcolor{blue!25}0.388 (0.181, 0.594)}                        & \cellcolor{blue!25}0.000                          & \multicolumn{1}{c|}{0.162 (-0.208, 0.533)}                         & 0.390                         \\ \hline
FST II, Age Over 65    & \multicolumn{1}{c|}{0.047 (-0.051, 0.144)}                        & 0.348                         & \multicolumn{1}{c|}{-0.003 (-0.094, 0.089)}                        & 0.955                          & \multicolumn{1}{c|}{-0.099 (-0.221, 0.023)}  & 0.112 \\ \hline
\end{tabular}
}}
\end{subtable}
\label{Tab:Diff_AUROC_model_Prove}

\end{table*}

Table~\ref{Tab:Diff_AUROC_model_Prove} depicts the differences in AUROC model discrimination between the risk models for all FST, sex and age combinations considered. Here we see multiple significant differences in terms of model performance. Starting with the comparisons of the top 3 models amongst themselves, ADAE has 4 significantly different results from 2nd Place (Everyone, FST 2 Men, FST 2 Women, FST 1 Age Over 65) and 5 from 3rd Place (Everyone, FST 2 Men, FST 1 Women, FST 2 Under 65, FST 2 Over 65). In all but 2 comparisons ADAE was the algorithm that had superior AUROC, highlighting that although ADAE did not perform as well on this dataset as on ISIC 2020, it still handles this new data better than its two closest competitors. Looking at the comparisons of the top 3 models to the baseline shows that all 3 models drop in performance and have multiple subgroups where their discrimination is not significantly better than the baseline; 3rd Place especially has performance that is not significantly better than ERM in any way. This contrasts what was seen when doing the same comparisons on the ISIC 2020 dataset, where each model was significantly better than the baseline amongst every subgroup, and suggests that 3rd Place having the highest AUROC was most likely due to overfitting to the ISIC data. 


In terms of AUROC discrimination, there was no significant difference in performance between the different subgroups of FST 1 and 2. In Table~\ref{Tab:Diff_AUROC_subgroup_Prove}, when stratified in terms of sex
we see AUROCs that range from 58\% (3rd Place, Women FST 1) to 93\% (2nd Place, Men FST 1) for FST 1, compared to 64\% (3rd Place, Women FST 2) and 85\% (ADAE, Men FST 2) for FST 2 (All P-values $>$ 0.05). 
When stratified in terms of age, we see AUROCs for FST 1 that range from 64\% (3rd  Place, FST 1 under 50) to 86\% (ADAE, FST 1 Over 50), compared to 47\% (3rd Place, Under 50 FST 2) and 83\% (ADAE, Over 50 FST 2) for FST 2 (All P-values $>$ 0.05). 
These results highlight the poor performance of the 3rd place model, which we explore further in the next section.



\subsection{Model Calibration}  \label{subsec: MC}

Table~\ref{Tab:Calibration} shows the calibration of various versions of the ADAE model using the Score-based CUSUM test of Feng et al. \textcolor{black}{We tested the calibration of the standard ADAE model (ADAE, Full Ensemble) and two randomly selected two EfficientNet models of the ADAE ensemble, one that uses metadata as a part of the inference process (EfficientNet-M) and one that does not (EfficientNet-NM). The inclusion of the two EfficientNet models was done to determine whether calibration was consistent across individual models in the ADAE ensemble or if it varied depending on architecture and metadata usage.}

\begin{table*}[t]
\centering
\caption{The results of Feng et al.'s calibration method on various versions of the ADAE model. Overall, it is shown across both datasets that ADAE is most likely miscalibrated in terms of age more than any other feature, meaning that if a person is to get a risk score too low or too high, their age plays a factor in this mis-assignment. The original risk prediction ('Prediction') is ignored in our analysis since the prediction is inherently tied to the outcome of this calibration experiment.}
\small
\begin{tabular}{|l|l|c|p{3.2cm}|c|p{3.2cm}|}
\hline
& & \multicolumn{2}{c|}{Underestimation} & \multicolumn{2}{c|}{Overestimation} \\
\hline
Model & Dataset & Test Statistic & VI Ranking & Test Statistic & VI Ranking \\
\hline
ADAE & ISIC 2020 & 8.45 & Prediction (1), Age (2) & 127.33 & Prediction (1), Age (2), Sex (3), Location: Athens (5) \\
\cline{2-6}
     & PROVE-AI  & 9.51 & Prediction (1), Age (2), Site: Lateral Torso (3), Sex (6) & 9.26 & Prediction (1), Age (2), Sex (3), Site: Lower Extremity (4) \\
\hline
EfficientNet-NM & ISIC 2020 & 14.04 & Prediction (1), Age (2), Location: Athens (3), Sex (4), Location: New York (5) & 21.15 & Prediction (1), Age (2), Sex (4) \\
\cline{2-6}
                & PROVE-AI  & 10.24 & Prediction (1), Age (2), Sex (3) & 10.53 & Prediction (1), Age (2), Site: Upper Extremity (3) \\
\hline
EfficientNet-M & ISIC 2020 & 13.63 & Prediction (1), Age (2), Location: New York (3), Sex (4) & 17.35 & Prediction (1), Location: Barcelona (2), Sex (4) \\
\cline{2-6}
               & PROVE-AI  & 11.18 & Prediction (1), Age (2), Sex (3) & 11.04 & Prediction (1), Age (2), Sex (3), Site: Lower Extremity (5) \\
\hline
\end{tabular}
\label{Tab:Calibration}
\end{table*}

As part of Feng et al.'s method, an ensemble of Kernel Logistic Regression residual models were trained on the metadata for each respective dataset to predict the true event rate for a given individual in that dataset. 
The eight residual models evaluated in this experiment varied in their hyperparameter configurations. The first four models used a regularization strength of $1 \times 10^{-3}$, with increasing values for the degree of the polynomial kernel approximation: Model 1 used degree 2, Model 2 used degree 3, Model 3 used degree 4, and Model 4 used degree 5. The remaining four models used a regularization strength of $1 \times 10^{-2}$, with the same structure: Model 5 used degree 2, Model 6 used degree 3, Model 7 used degree 4, and Model 8 used degree 5. All eight models shared a maximum iteration limit of 2000, used L2 regularization to prevent overfitting, and applied no weighting to zero-labeled samples. Because ADAE was trained previously, we dedicated all data in each respective dataset to training the residual models and testing for strong calibration. \textcolor{black}{Input features to the residual models included demographic information that were available for each dataset (Age and Sex for ISIC; Age, Sex and FST for PROVE-AI) as well as the location of the lesion on the body (Site) and the hospital that the image was collected at (Location).} To increase the accuracy of the residual models, we also extracted intermediate features from each model being tested and used those as additional features to train the ensemble of residual models; in the figures below, these intermediate features are labeled as \emph{Feature Embedding (Number)}. \textcolor{black}{We used the \emph{CVScore} variation of this test which performs cross validation by training residual models on subsets of the data and testing for calibration on held-out folds.}
Our tolerance level for calibration, denoted as $\delta$ in the original paper, was set to 0 for all experiments. We provide the technical details of this method in Section~\ref{sec:Methods}.

\textcolor{black}{This test computes P-values using a Monte Carlo approach by simulating the distribution of a CUSUM-based test statistic under the null hypothesis of perfect calibration. The P-value is then estimated as the proportion of simulated test statistics that exceed the observed test statistic.
For each experimental result discussed below, the P-value was 0.0 so we reject each null hypothesis and assume there’s at least one, but possibly multiple, subgroups that are being either Overestimated (i.e. they have a true risk lower than their predicted value) or Underestimated (i.e. they have a true risk higher than their predicted value).}

\subsubsection{ISIC 2020}

\begin{figure*}[h]
\centering
\begin{subfigure}{\textwidth}
    \centering
  \includegraphics[width=\linewidth]{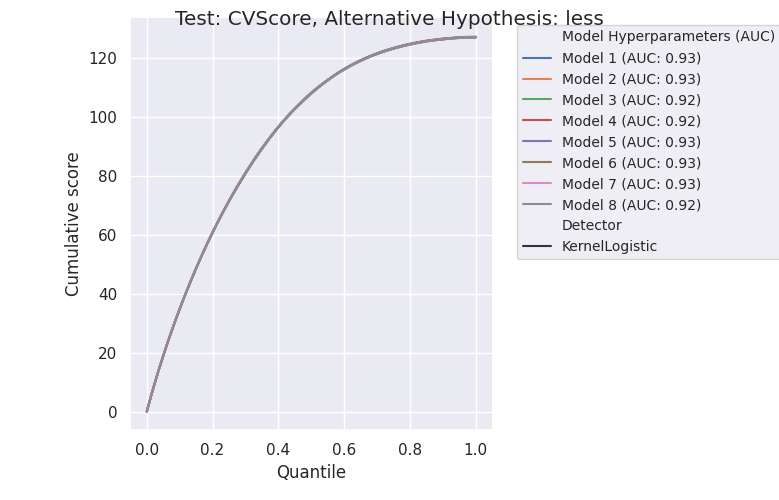}
  \caption{A control chart plotting cumulative score. Here, the ADAE model is being checked for overestimation on the ISIC 2020 dataset.}
  \label{fig:isic_plot7}
\end{subfigure}
\begin{subfigure}{0.7\textwidth}
  \centering
  \includegraphics[width=\linewidth]{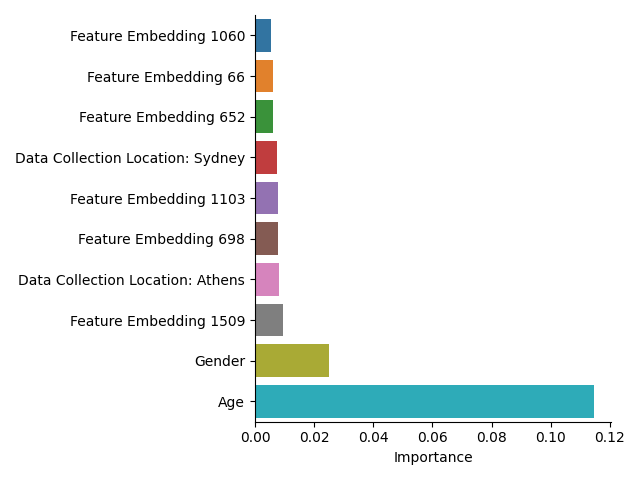}
  \caption{Variable Importance Plot for the top 10 features.}
  \label{fig:isic_feat_import7}
\end{subfigure}%
\caption{A control chart and Variable Importance plot to check for subgroups who have their true risk overestimated in the ISIC 2020 dataset.}
\label{fig:isic_overestimation}
\end{figure*}

\begin{figure*}[h]
\centering
\begin{subfigure}{\textwidth}
    \centering
  \includegraphics[width=\linewidth]{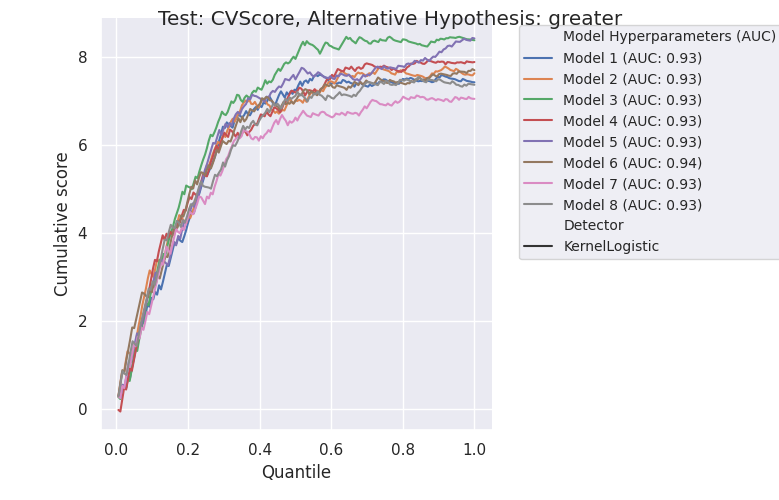}
  \caption{A control chart plotting cumulative score. Here, the ADAE model is being checked for underestimation on the ISIC 2020 dataset.}
  \label{fig:isic_plot5}
\end{subfigure}
\begin{subfigure}{0.7\textwidth}
  \centering
  \includegraphics[width=\linewidth]{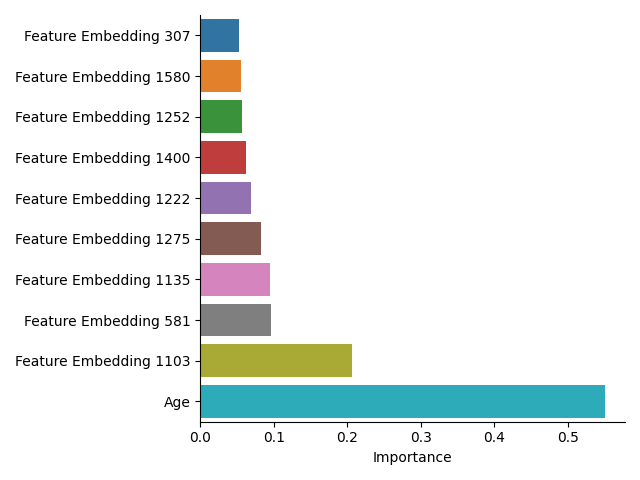}
  \caption{Variable Importance Plot for the top 10 features.}
  \label{fig:isic_feat_import5}
\end{subfigure}%
\caption{A Control Chart and Variable Importance plot to check for subgroups who have their true risk underestimated in the ISIC 2020 dataset.}
\label{fig:isic_underestimation}
\end{figure*}



Figures~\ref{fig:isic_overestimation} and~\ref{fig:isic_underestimation} respectively show the tests for overestimation and underestimation in the ADAE model on the ISIC 2020 dataset. 
The most important features are shown in ascending order on the y-axis of the Variable Importance plots (Figures~\ref{fig:isic_feat_import7} and~\ref{fig:isic_feat_import5}). The importance of that feature is defined as the drop in the test statistic after randomly permutating the feature's value, where a larger drop indicates a more important feature.

\textcolor{black}{Focusing first on overestimation, we see the test statistic in Figure~\ref{fig:isic_plot7} quickly rise to a value multiple times higher than that of the one seen in Figure~\ref{fig:isic_plot5} (127.33 vs. 8.45, respectively). This difference in value is due to the difference in the number of samples used in each test; The overestimation case only uses samples that produce positive predicted residuals to build the CUSUM Plot in Figure~\ref{fig:isic_plot7}, while the underestimation case uses samples with negative residuals to do the same. These results suggest that, in the ISIC 2020 dataset, the ADAE model is more prone to overestimating risk than underestimating it.}

The case for ADAE overestimating a large number of samples is further supported by the high number of False Positives that we observed in Table~\ref{Tab:95_pct_ISIC}. 
For underestimation, we see in Figure~\ref{fig:isic_plot5} that there is an sharp rise in the test statistic across all residual models, followed by a slower increase. These results suggest that there is at least one subgroup identified by the kernel logistic model that is poorly calibrated. 
The VI plots in Figures~\ref{fig:isic_feat_import7} and~\ref{fig:isic_feat_import5} further suggest that the most important features that characterize this poorly calibrated group are age and the original risk prediction. 
The same features were listed as the most important when running this method on EfficientNet-NM. In the EfficientNet-M model however, age did not show as a strong factor for overestimation, suggesting that the inclusion of patient metadata in the training process may help with controlling over-diagnosis due to age. Interestingly, Both EfficientNet-M and EfficientNet-NM  listed various Locations as a high cause for miscalibration while the ADAE model did not list a location in the top 3 features. This difference in the effect of location may be specific to these ResNet models, but not to the other models that make up the ensemble.

\subsubsection{PROVE-AI}

\begin{figure*}[h]
\centering
\begin{subfigure}{\textwidth}
    \centering
  \includegraphics[width=\linewidth]{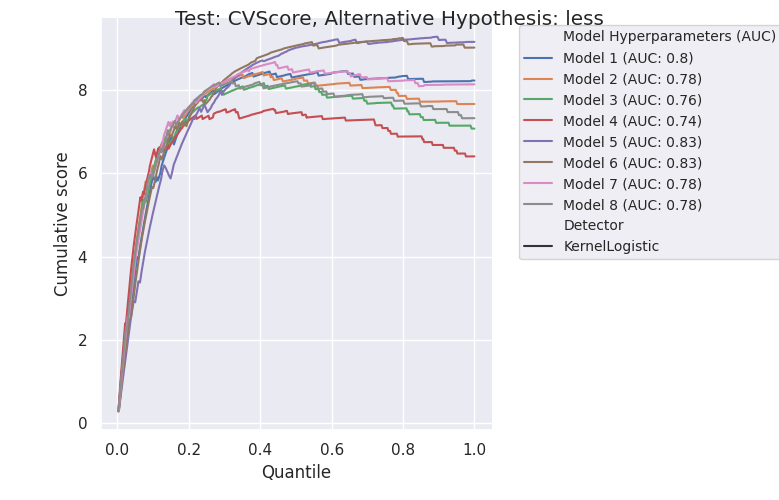}
  \caption{A control chart plotting cumulative score. Here, the ADAE model is being checked for overestimation on the PROVE-AI dataset.}
  \label{fig:prove_plot3}
\end{subfigure}
\begin{subfigure}{0.7\textwidth}
  \centering
  \includegraphics[width=\linewidth]{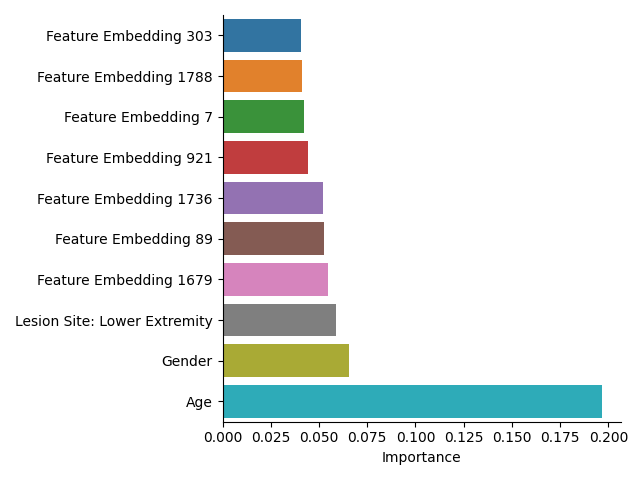}
  \caption{Variable Importance Plot for the top 10 features.}
  \label{fig:prove_feat_import3}
\end{subfigure}%
\caption{A control chart and variable importance plot to check for subgroups who have their true risk overestimated in the PROVE-AI dataset.}
\label{fig:prove_overestimation}
\end{figure*}

\begin{figure*}[h]
\centering
\begin{subfigure}{\textwidth}
    \centering
  \includegraphics[width=\linewidth]{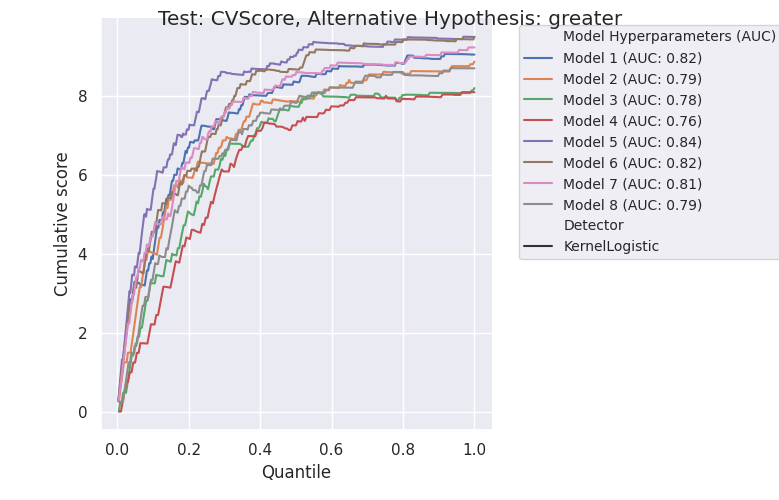}
  \caption{A control chart plotting cumulative score. Here, the ADAE model is being checked for underestimation on the PROVE-AI dataset.}
  \label{fig:prove_plot1}
\end{subfigure}
\begin{subfigure}{0.7\textwidth}
  \centering
  \includegraphics[width=\linewidth]{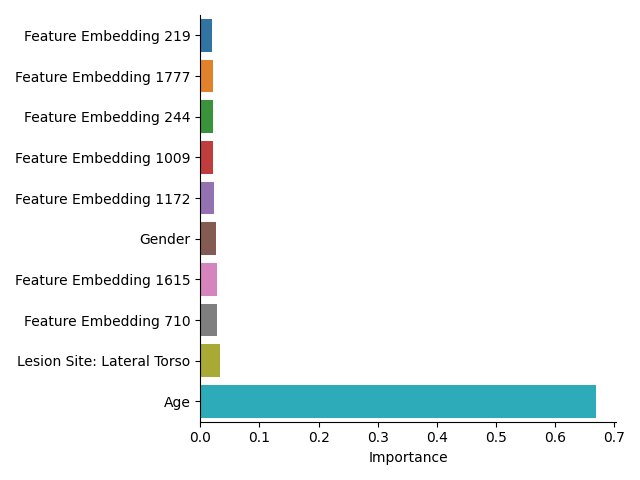}
  \caption{Variable Importance Plot for the top 10 features.}
  \label{fig:prove_feat_import1}
\end{subfigure}%
\caption{A control chart and variable importance plot to check for subgroups who have their true risk underestimated in the PROVE-AI dataset.}
\label{fig:prove_underestimation}
\end{figure*}


Figures~\ref{fig:prove_overestimation} and~\ref{fig:prove_underestimation} respectively show the tests for overestimation and underestimation in the ADAE model on the PROVE-AI dataset. Similar to what we saw with ADAE in our Model Discrimination analysis, the AUROC of each residual model in the ensemble decreased. 
Looking at both figures, we see the test statistic quickly rise but then smoothen out, suggesting again that there is one subgroup identified by the kernel logistic model that is poorly calibrated. The VI plots in Figures~\ref{fig:prove_feat_import3} and~\ref{fig:prove_feat_import1} further suggest that the most important features that characterize this poorly calibrated group are age and the original risk prediction. 
The same features were listed as the most important when running this method on EfficientNet-NM and EfficientNet-M. All 3 models also on average listed Sex as a higher cause for miscalibration in this dataset than in ISIC 2020. 
Somewhat surprisingly, we do not see FST as 
a strong factor of (mis)calibration.

\section{Discussion and Conclusion} \label{sec: DC}

\subsection{Summary of Results}

Existing AI models tended to significantly improve discriminative accuracy for melanoma compared to a baseline Expected Risk Minimization (ERM) algorithm when applied to two separate datasets, one from the ISIC 2020 Challenge, where the models were developed, and the PROVE-AI dataset, which featured risk factors that were not present in the former dataset. Prior models generally performed worse on the PROVE-AI dataset, sometimes to the level of ERM, with the best performance exhibited by the ADAE model, which uses an ensemble of 90 models to make predictions. 

The model discrimination of ADAE, developed on data from 3 previous ISIC challenges, was significantly better than the other 2 models, developed on data from 2 previous challenges, when applied to the PROVE-AI dataset. Due to differences in the distribution of each dataset, a decrease in performance as observed from the ISIC 2020 test dataset to prospective studies, is expected. Another potential reason for the difference in performance could be the inclusion of patient risk factors in PROVE-AI that was not present in ISIC 2020, such as FST. 
Another possible explanation might be an inadequate ability of existing models to generalize to unseen data. Focusing on this last issue and exploring the extent to which AI models can improve model performance across sex, FST, and age groups, a baseline AI model, ERM, was trained and validated using data from a previous ISIC Challenge. Overall, ERM had better discriminative performance on ISIC 2020 and experienced a similar drop in performance to the other models when applied to the PROVE-AI dataset. When compared directly, ERM sometimes offered even better performance than 2nd Place and 3rd Place. These findings are consistent with other studies suggesting that there can be limited benefit in using more complex bias mitigation strategies over ERM~\citep{zong2023medfairbenchmarkingfairnessmedical}. 

Poor discrimination was largely driven by the results in the PROVE-AI dataset. Although event rates vary depending on an individual's age, sex and FST, the ability of ADAE to risk rank individuals was significantly weaker for FST 1 participants than for FST 2 participants for both sexes, and for both age groups. This inaccurate risk assignment for individuals has the potential for ineffective intervention, where higher-risk individuals do not receive beneficial therapy and lower-risk individuals are over-treated. \textcolor{black}{However, it is important to note this discriminative analysis of FST is not complete due to the lack of available FST data for types 5 and 6. Due to this, we refrain from making a strong statement on the discriminative ability of ADAE on the basis of race and instead leave this form of analysis on a larger, more inclusive dataset as future work.} 

Using the calibration method of Feng et al., we see that ADAE was surprisingly prone to overestimation on ISIC 2020. ADAE also was calibrated well in terms of FST on the PROVE-AI dataset. When examining two of the ResNet-18 models that make up this ensemble, it was shown that different predictive factors led to overestimation and underestimation in the two models. This shows that the inclusion or exclusion of patient metadata in the training of an AI model does indeed play a role in the final calibration of the model, although more analysis needs to be done to determine whether this role is overall beneficial or harmful to the final result. Future work may look to repeat this study with varying combinations of metadata to gain a better understanding of what factors contribute the most to model discrimination.

In terms of FST, more significant differences were observed in terms of model discrimination than calibration. \textcolor{black}{This may be due to limited representation of darker skin tones in the dataset, the relatively lower influence of FST on predicted probabilities compared to features like age and sex, or the calibration method's sensitivity favoring features with stronger statistical signals or larger subgroup sizes.} Each form of analysis can be caused by similar issues during the model's creation but can lead to different consequences. Observing multiple differences in model discrimination and calibration implies that the subgroups included in the model training not only require different amounts of information to obtain accurate predictions, but also that the way in which the model learns to predict based on the information provided may favor one subgroup over another. In terms of model discrimination, these two factors could lead to differences in a model's ability to distinguish between positive and negative samples for a subgroup; an example of this would be the difference in AUC observed for FST 1 and 2 for 2nd Place in Table~\ref{Tab:Diff_AUROC_subgroup_ISIC}, which shows that 2nd Place is significantly better at distinguishing benign cases from malignant for FST 1 than it is for FST 2. 
In terms of model calibration however, these two factors could lead to unequal preventive measures being applied to a population. As shown by the example in Section~\ref{sec:Intro}, overestimation of risk can lead to unnecessary over-treatment; additionally, underestimation of risk can lead to reduced access to care.   

\textcolor{black}{Although the best performing models in our study, particularly ADAE, demonstrate strong performance, they also rely on large ensembles of deep neural networks, which require substantial computational resources for training and inference. In contrast, the ERM baseline, which consists of a single ResNet model trained without metadata, requires significantly fewer resources and achieves competitive performance in certain settings, especially on the PROVE-AI dataset. This raises important questions about the tradeoff between computational cost and model performance. In clinical settings, where real-time decision-making, limited hardware availability, and cost constraints are common, models that are computationally efficient may be more practical—even if they offer slightly lower performance. Additionally, simpler models are often easier to validate, interpret, and maintain, which is critical for regulatory approval and clinical trust. Future work should explore this tradeoff more explicitly by reporting metrics such as training time, inference latency, model size, and energy consumption. Such analysis would be valuable for assessing the feasibility of deploying these models in real-world healthcare environments, particularly in low-resource or point-of-care settings.}

Despite some of the results showing worse performance, we do not ignore the potential that AI models provide in this domain. Deep learning models have been highly effective in risk prediction because they can extract highly complex, latent features in high-dimensional data sets. As stated in the previous section, future work should look to understand the effect that training a Deep Learning model with different combinations of demographic 
(race, sex, and age) and clinical (location, lesion type) information has on its performance in our analysis. 
Other potential solutions to the performance issues outlined in this section could be a technique such as transfer learning~\citep{zhuang2020comprehensivesurveytransferlearning}, where an AI model trained using data from one distribution of people would be adapted to a second population with a different distribution of demographics and clinical features. Additionally, a technique such as federated learning could be used to train multiple models at once with data from multiple populations~\citep{Kaur2023}. 

\subsection{Limitations}


This study was designed to benchmark new and existing techniques for comparing melanoma detection models. In this section, we acknowledge some limitations.

\textcolor{black}{First, we used Fitzpatrick Skin Type (FST) as a proxy for skin tone in evaluating model fairness. However, we acknowledge that FST is not equivalent to race. FST is a dermatological classification based on skin’s response to ultraviolet radiation and does not always capture the broader social, cultural, or systemic dimensions associated with racial identity. Therefore, we caution against interpreting our subgroup analyses as comprehensive assessments of racial fairness. Future work should incorporate more nuanced and demographic data to better understand and address racial disparities in dermatologic AI systems.}

Additionally, there was little to no FST data available for 4 of the 6 categories. This prevented us from performing more comparisons of AUC, and therefore limited the experimental results we obtained in terms of FST. Despite this, the analysis we did with FST 1 and 2 did provide an avenue for future work that can be explored further once a more balanced dataset is available. 

Along this same line of limitations, only two datasets were used in this experiment. While this is equivalent to the number of skin cancer datasets that were used in similar benchmarks~\citep{zong2023medfairbenchmarkingfairnessmedical}, this again highlights how future work along this line can redo an experiment of this type with more data and datasets in order to better understand the performance of these top ranking models.

\textcolor{black}{Another limitation lies in the heterogeneity of the evaluated models. The top three ISIC 2020 entries differ not only in ensemble size but also in underlying architecture (e.g., EfficientNet vs. ResNet) and metadata usage. These differences introduce confounding variables that make it difficult to isolate the specific factors contributing to performance or calibration differences. A more controlled comparison—e.g., training a single architecture under varying conditions (with/without metadata, single model vs. ensemble)—would help disentangle these effects. While such an analysis may be beyond the scope of this study, acknowledging this variability is important for interpreting the results.}

\textcolor{black}{Furthermore, the models evaluated in this study are large ensembles of CNNs trained across shared cross-validation folds, each with varying parameters. In particular, the ADAE model evaluated in this study is an ensemble of 90 models (Eighteen CNNs each with Five-fold cross validation), with and without metadata. While this ensembling strategy likely contributes to stronger performance, it also introduces complexity that may obscure the behavior of individual models and complicate interpretability. The shared cross-validation setup may limit reproducibility and hinder a more granular understanding of how fairness manifests across the ensemble. We address this by examining two models of this ensemble in our calibration experiments, but we acknowledge that future work could benefit from evaluating simpler or more transparent base models to better isolate the effects of discrimination and calibration techniques.}

Finally, the limitations of the calibration method used: the method assumes that residuals can be accurately predicted and that changepoints in residuals correspond to poorly calibrated subgroups. If these assumptions do not hold, the method's reliability may be affected. Future work could look at measuring the uncertainty of the predicted residuals to ensure that the final results are as accurate as possible.

\textcolor{black}{Additionally, while this calibration method (based on the CUSUM test and Variable Importance plots) is designed to detect miscalibration across combinations of demographic subgroups, our results consistently identified only one subgroup as miscalibrated per test. This may reflect limitations in the method’s sensitivity to intersectional effects, especially when subgroup sizes are small or when signal strength is diluted across overlapping identities. Additionally, the datasets used in this study may lack sufficient representation of certain intersectional subgroups (e.g., older women with darker skin tones), which could hinder the detection of nuanced calibration disparities. Future work should explore calibration methods with enhanced sensitivity to intersectional miscalibration and apply them to larger, more demographically diverse datasets to better capture these effects.}

\subsection{Conclusion}


While AUROC-based discrimination measures are the more popular way of comparing a model's performance to another, evaluating the calibration of its predictions is also important, especially in clinical applications. 

We conducted an extensive benchmarking of existing melanoma detection models to better assess subgroup performance in relation to dataset composition and subgroup size. By evaluating Model Discrimination and Model Calibration metrics, we identified key discrepancies and limitations in the top 3 algorithms for the ISIC 2020 Challenge with respect to fairness, demonstrating the promise of calibration as an auxiliary fairness metric for dermatology algorithms.

Our findings indicate that existing models sometimes perform to the level of a baseline model when applied to datasets with previously unseen risk factors. 
Additionally, the inclusion or exclusion of patient risk factors causes discrepancies in the final performance of a model, although more information is needed to determine exactly in what way.
These results highlight the continued need for more extensive metadata collection from subgroups of interest to achieve dermatologist-level classification performance in this task. 
Overall, this study serves as a guideline for additional development and auditing of melanoma detection models, as well as model discrimination and calibration methods that can comprehensively evaluate them.

\section{Acknowledgments}

Funding for this work was provided by NIH/NCI grants U24-CA285296, U24-CA264369, R01CA293974; United States Department of Defense grants HT94252410552, HT94252410553, HT94252410554, HT9425-23-1-0848; NIH/NCI Cancer Center Support grant P30 CA008748; and the Burroughs Welcome Fund Innovation in Regulatory Science Award.




\bibliography{sample}

\end{document}